\documentclass{article}
\usepackage{iclr2026_conference,times}

\usepackage{hyperref}
\usepackage{url}
\usepackage{times}
\usepackage{latexsym}
\usepackage{amsmath}
\usepackage{amssymb}
\usepackage{booktabs}
\usepackage{graphicx}
\usepackage{subcaption}
\usepackage{enumitem}
\usepackage[T1]{fontenc}
\usepackage[utf8]{inputenc}
\usepackage{microtype}
\usepackage{xcolor}

\iclrfinalcopy

\title{SpecOpt: Contact-Diff Reasoning for Agentic Molecule Optimization Toward Binding Specificity}

\author{Thao Nguyen \& Heng Ji \\
Siebel School of Computing and Data Science \\
University of Illinois Urbana-Champaign \\
\texttt{\{thaotn2, hengji\}@illinois.edu}}

\begin{document}
\raggedbottom
\maketitle

\begin{abstract}
Off-target protein binding is a major source of adverse effects for small-molecule drugs, yet most structure-based molecular design methods focus on generating selective compounds \emph{de novo} rather than improving the selectivity of existing, well-characterized drugs. We introduce \emph{specificity optimization} (SpecOpt), a molecular design task that seeks constrained structural modifications to an existing compound that increase its binding preference for an intended target over known off-targets while preserving its structural identity and drug-like properties. To enable systematic evaluation of this task, we construct a ChEMBL-derived benchmark from compound--target interaction data, identifying intended targets through curated drug-mechanism annotations and off-targets through measured activities. We then develop an agentic framework for SpecOpt that docks each compound against its intended target and off-targets, compares the resulting poses through residue-aware atom--protein contacts, and provides these differential interactions to a large language model to propose targeted structural modifications. Candidates are retained only if they satisfy molecular similarity, ADMET, and target--off-target docking selectivity criteria. On 915 compounds, the agent improves the target--off-target binding gap for 84.8\% of compounds, shifting the mean gap from $-0.72$ to $+0.47$~kcal/mol while maintaining a mean Tanimoto similarity of $0.72$ to the starting compounds. Ablation studies identify residue-specific contact information as the critical optimization signal: replacing residue identities with binary contact indicators eliminates improvement on all 29/29 ablation compounds. These results establish SpecOpt as a distinct molecular design problem and demonstrate residue-aware differential interactions as an effective signal for improving the specificity of existing compounds.
\end{abstract}

\section{Introduction}

Small-molecule drugs rarely interact exclusively with their intended targets. Polypharmacology is widespread~\citep{keiser2007rea}, and unintended interactions with off-target proteins can contribute to adverse effects, dose-limiting toxicity, and clinical failure. Improving molecular selectivity is therefore a fundamental objective in drug development. Recent computational approaches have increasingly addressed selectivity through structure-based generative design, seeking to generate new molecules that bind strongly to a desired protein while avoiding alternative targets~\citep{chandra2024bayesian, southiratn2025combimots, theselective2026}. These methods address an important design problem, but they largely formulate selectivity as a property to optimize during \emph{de novo} molecular generation.

A different problem arises once a promising drug or drug candidate already exists. Such a compound may have favorable potency, pharmacokinetic properties, synthetic accessibility, extensive experimental characterization, or even established clinical use, while retaining undesirable activity against one or more off-targets~\citep{lounkine2012offtargets}. In this setting, replacing the molecule with a newly generated compound is not necessarily the desired solution. Instead, our objective is to modify the existing molecule as little as possible while selectively weakening its off-target interactions and preserving the properties that made it a viable drug candidate. This goal builds on evidence that structural modifications can alter off-target activity~\citep{wassermann2010selectivity} and on medicinal chemistry principles of addressing liabilities while maintaining on-target potency and favorable physicochemical properties~\citep{quancard2025hittolead}. We refer to this task as \emph{specificity optimization (SpecOpt)}: given a starting compound, an intended target, and one or more known off-targets, identify constrained structural modifications that increase the binding preference for the intended target while preserving the identity and drug-like characteristics of the starting molecule.

To our knowledge, specificity optimization of an existing drug has not previously been formulated and systematically studied as a standalone computational molecular design task. This formulation differs fundamentally from \emph{de novo} selective generation~\citep{chandra2024bayesian}. The starting molecule is fixed rather than generated, the permissible structural deviation is constrained, and the optimization objective is inherently comparative: success requires increasing the \emph{specificity gap} between binding to the intended target and binding to off-target proteins. Consequently, a successful method must determine not only which molecular structures are compatible with the target, but which modifications to an existing structure can preferentially alter its interactions with one protein over another.

The fixed starting molecule also makes available a direct structural signal for solving this problem. The same compound can be docked against its intended target and off-targets, allowing its binding modes to be compared atom by atom. Rather than asking only whether a molecular feature contributes to binding, we ask a differential question: \emph{how does each part of the molecule interact with the intended target differently from the off-targets?} This comparison can identify regions that preferentially support target or off-target binding and, importantly, regions that occupy similar positions across binding pockets but interact with different residues. The latter distinction is especially relevant for closely related proteins, where a ligand may contact nearly the same regions of homologous pockets even though the identities and chemical environments of those contacts differ.

\begin{figure}[t]
\centering
\includegraphics[width=\textwidth]{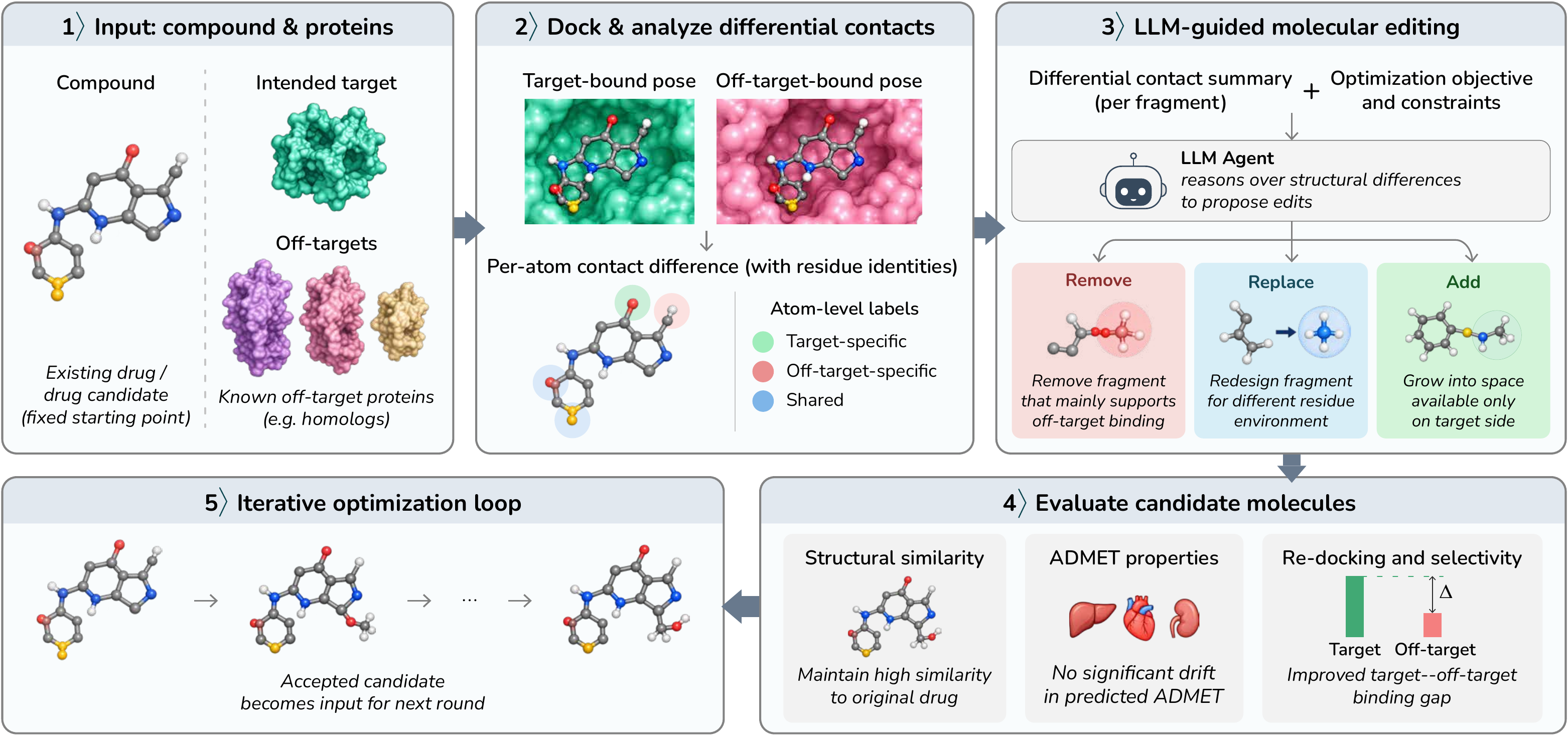}
\caption{\textbf{Overview of SpecOpt.} Given a compound, its intended target, and known off-targets, SpecOpt docks the compound against each protein, computes per-atom contact differences with residue identities, and uses an LLM agent to propose targeted molecular edits. Candidate molecules are evaluated using structural similarity, ADMET, and re-docking criteria, and accepted candidates are iteratively refined in subsequent rounds.}
\label{fig:architecture}
\end{figure}

We introduce an agentic framework for this specificity-optimization task (Figure~\ref{fig:architecture}). Given a compound, its intended target, and known off-targets, the framework docks the compound against each protein and constructs a residue-aware, per-atom representation of the differences between the resulting binding poses. These signals are aggregated over molecular fragments and provided to a large language model (LLM) agent, which reasons over the differential interaction environment to propose targeted structural edits. The agent can remove fragments that preferentially support off-target interactions, modify regions exposed to different residue environments, or extend the molecule toward target-specific interaction opportunities. Proposed molecules are then evaluated using structural similarity, ADMET, and re-docking criteria, ensuring that optimization remains anchored to the original compound while improving its predicted specificity.

Our main contributions are:

\begin{itemize}[leftmargin=1.4em,itemsep=1pt,topsep=2pt]

\item \textbf{We formulate specificity optimization as a computational molecular design task.} To our knowledge, this is the first framework to explicitly address the problem of modifying an existing drug or drug candidate to increase its target--off-target binding preference while constraining structural and drug-like-property changes. This task complements, rather than replaces, existing work on selective \emph{de novo} molecular generation.

\item \textbf{We introduce a benchmark and agentic framework for the task.} We construct a ChEMBL-derived dataset of 1{,}848 compounds annotated with intended targets and experimentally measured off-targets and develop an iterative agent combining differential structural analysis, LLM-guided molecular editing, and property-based filtering. On 915 evaluated compounds, the framework improves the docking-predicted target--off-target specificity gap for 84.8\% of compounds, while maintaining a mean Tanimoto similarity of 0.72 to the starting molecules.

\item \textbf{We identify residue-aware differential contacts as the key optimization signal.} Successful optimization depends not merely on whether ligand atoms contact the target or off-target, but on \emph{which residues} they contact. Replacing residue-aware contact differences with a simpler target/off-target/shared classification eliminates improvement on all 29/29 compounds in our ablation study, highlighting the importance of residue identity for distinguishing interactions across similar binding pockets.

\item \textbf{We systematically evaluate additional agentic mechanisms.} We investigate cross-compound memory, pose-consistency caution, and block-synergy mechanisms designed to provide additional experience and structural context. Although these mechanisms produce their intended intermediate behaviors, they do not measurably improve final optimization outcomes, helping isolate which information is consequential for this task.

\end{itemize}

\section{Related Work}

\paragraph{Selectivity-aware molecular generation.} Recent structure-based generative methods have incorporated selectivity by optimizing binding to an intended target while penalizing activity against off-targets. \citet{chandra2024bayesian} use Bayesian optimization over the latent space of a junction-tree VAE~\citep{jin2018junction}, rewarding docking to FLT3 while penalizing docking to homologous kinases. \citet{theselective2026} steer molecular generation with dual affinity guidance, while \citet{southiratn2025combimots} use Pareto Monte Carlo tree search over fragment space for the related but inverse objective of dual-target binding. These approaches construct molecules \emph{de novo}, whereas SpecOpt begins with an existing compound and seeks constrained structural modifications that improve its binding preference for the intended target while preserving similarity to the original molecule. This setting also enables direct comparison of how the same molecule interacts with its target and off-targets. We compare with \citet{chandra2024bayesian} on FLT3 in Section~\ref{sec:flt3}, while noting that absolute selectivity is not directly comparable because the two methods address different molecular design settings.
\vspace{-0.8em}

\paragraph{Structure-based lead optimization.} Several structure-based methods modify existing ligands using information from the protein binding pocket. DeepFrag~\citep{green2021deepfrag} predicts fragment additions from a receptor--ligand complex, while STRIFE~\citep{hadfield2022strife} uses target-derived pharmacophoric information to guide fragment elaboration. AutoGrow4~\citep{spiegel2020autogrow4} applies docking-guided genetic search to evolve existing ligands toward improved predicted binding, and FRAME~\citep{powers2023geometric} uses geometric deep learning to iteratively grow fragments from a bound ligand, reporting improvements in both predicted affinity and selectivity. These methods demonstrate that protein structural context can effectively guide modification of an existing molecular starting point. SpecOpt differs in explicitly reasoning over the \emph{difference} between how the same molecule interacts with its intended target and known off-targets, and using these differential interactions to guide constrained molecular edits. To our knowledge, SpecOpt is the first framework to formulate specificity optimization of an existing compound against known off-targets as a standalone molecular design task.

\vspace{-0.8em}
\paragraph{LLM agents for molecular design.} Tool-augmented LLMs have been applied broadly to chemistry~\citep{bran2024chemcrow} and more recently to lead optimization. \citet{li2026trace} formulate tool selection as sequential decision-making over molecular edit trajectories, while \citet{wang2026molmem} incorporate memory into an agentic reinforcement learning optimizer. MolLingo~\citep{nguyen2026mollingo} further demonstrates that LLMs can reason over chemically meaningful molecular fragments together with docking-derived residue and binding-pocket context to propose molecular modifications that improve predicted target binding. Building on this capability, SpecOpt focuses specifically on target-versus-off-target specificity, providing the LLM with differential structural information between target- and off-target-bound poses to guide molecular editing.

\section{SpecOpt: Contact-Diff-Guided Specificity Optimization}

\subsection{Task Formulation}

Given a compound $m$, an intended target $T$, and a set of off-targets $\mathcal{O}$, let $s(m, P)$ be the AutoDock Vina~\citep{trott2010autodock} score of $m$ against protein $P$ (more negative = tighter binding). We define the \textbf{specificity gap}
\begin{equation}
g(m) \;=\; \min_{O \in \mathcal{O}} s(m, O) \;-\; s(m, T),
\label{eq:gap}
\end{equation}
the margin between the target and the \emph{tightest-binding} off-target. A positive gap means the molecule prefers its target over every off-target; the optimization objective is to increase $g$ relative to the original compound $m_0$, subject to staying chemically close to $m_0$.

\subsection{Benchmark construction}

We derive a specificity benchmark from ChEMBL~\citep{mendez2019chembl}'s compound--target interaction data. A compound's \emph{intended} targets are those with a curated drug-mechanism designation; its \emph{off-targets} are proteins with measured activity but no such designation. Constructing this cleanly required correcting three systematic over-counting problems: mutation panels, where different point mutations of the same protein are recorded as separate targets but are collapsed here because we focus on cross-protein specificity rather than selectivity among protein variants; protein complexes and subunits, where a subunit is counted separately from its parent complex; and generic-organism duplicates, where the same protein is registered under both a specific and an unspecified organism. The resulting dataset contains \textbf{1{,}848 compounds}, with a median of 3 and a mean of 7.5 off-targets each.

For docking-based experiments we filter to compounds where both the intended target and at least one off-target are single proteins with a resolvable experimental or predicted structure, and additionally exclude conformationally extreme molecules ($>45$ heavy atoms or $>10$ rotatable bonds), since docking cost scales with ligand flexibility. This yields a working pool of \textbf{916 compounds}. To make repeated experimentation tractable we precompute baseline docking for all 15{,}195 dockable (compound, target) pairs in the dataset, of which \textbf{13{,}636} (89.7\%) resolve successfully.

\subsection{Differential interaction analysis}
\label{sec:signal}

The central signal in SpecOpt comes from comparing how the same ligand interacts with its intended target and off-targets. For each ligand atom $i$, we identify protein residues within 7~\AA{} in the target-bound pose and each off-target-bound pose. Based on these contacts, atoms are classified as \textsc{target-only}, \textsc{offtarget-only}, \textsc{shared-same-residues}, or \textsc{shared-different-residues}. The last two categories distinguish atoms that contact the same residue identities across proteins from those that occupy similar regions of the binding pockets but interact with different residues. Independently, an atom is marked as a \textsc{growth-opportunity} when there is measurably more unoccupied pocket volume near that atom in the target than in the off-target.

We aggregate these atom-level signals over BRICS~\citep{degen2008brics} fragments and assign each fragment an editing suggestion: \textsc{remove}, \textsc{keep}, \textsc{redesign}, \textsc{grow}, or \textsc{neutral}. These suggestions determine how each fragment and its local interaction environment are presented to GPT-5.4~\citep{openai2026gpt54}, which serves as our LLM agent. Importantly, contact with both proteins does not imply that an atom is uninformative: when the contacted residues differ between the target and off-target, the position provides an opportunity for a chemical modification to preferentially affect one interaction environment. As shown in Section~\ref{sec:ablation}, preserving this residue-level distinction is critical to the effectiveness of SpecOpt.

\begin{figure}[htbp]
\centering
\includegraphics[width=\textwidth]{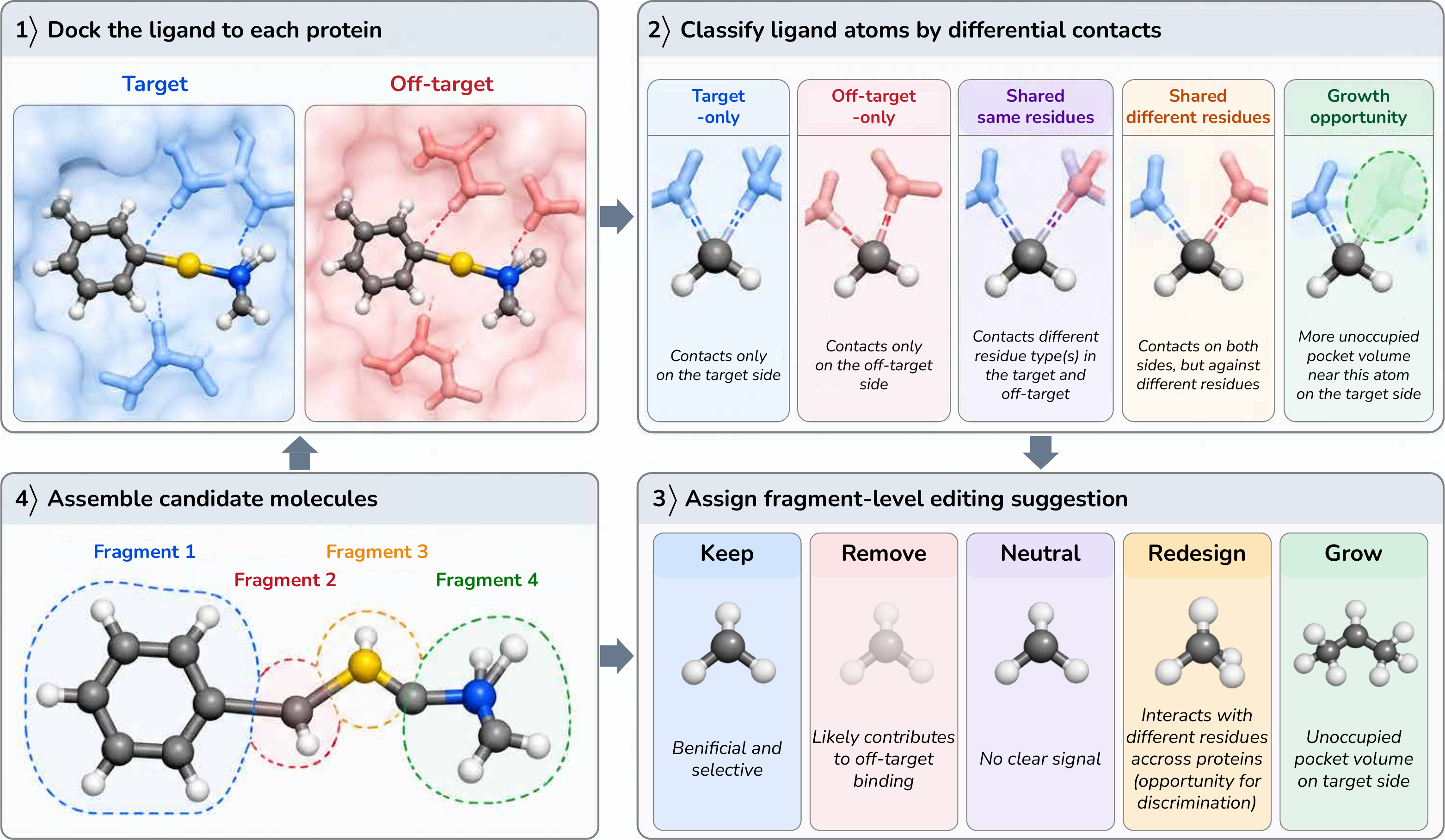}
\caption{\textbf{Differential interaction analysis and molecular editing in SpecOpt.} Target- and off-target-bound poses provide atom-level contact differences and growth opportunities, which are aggregated into fragment-level editing suggestions to guide candidate molecule construction.}
\label{fig:method}
\end{figure}

\subsection{Iterative optimization and candidate filtering}

Starting from the original compound, SpecOpt performs iterative optimization by alternating between LLM-guided molecular editing and candidate evaluation. At each round, the fragment-level interaction summary, current docking scores, and previously attempted edits are provided to the LLM, which proposes fragment-level modifications. These proposals are applied to generate candidate molecules, with one fragment modified at a time. Each candidate must pass three criteria before acceptance: Tanimoto similarity to the \emph{original} molecule above 0.4, ADMET changes within tolerance for direction-unambiguous toxicity endpoints (AMES, DILI, hERG, and P-gp), and an improved target--off-target specificity gap after re-docking. We evaluate ADMET properties using the prediction model from mCLM~\citep{edwards2026mclm}. Accepted candidates become the starting molecules for the next round, allowing structural modifications to accumulate over successive iterations. We additionally evaluate two extensions in Section~\ref{sec:ablation}: a \emph{combo} mechanism that combines independently beneficial edits, and a two-level memory mechanism consisting of within-compound edit history and cross-compound memory retrieval.

Because both docking and LLM generation introduce stochasticity, we fix the RDKit~\citep{landrum_rdkit} conformer seed and Vina search seed throughout the experiments. Fixing both is necessary for reproducible docking scores; otherwise, identical inputs can produce slightly different scores (e.g., $-7.1$ vs.\ $-7.2$~kcal/mol), which can in turn change which edits are selected. LLM proposals retain residual stochasticity, which defines the noise floor against which the ablation results are interpreted.

\section{Experiments}

\subsection{Main result}

We run the full system over the 916-compound pool with a budget of 8 candidate proposals per round and 4 rounds per compound. Failed runs are retried up to two additional times, for a maximum of three attempts per compound; a compound is excluded if all three attempts fail. Across 945 attempts, including retries, 915 distinct compounds complete successfully, while one compound exhausts the retry limit. The 30 failed attempts (3.2\% of all attempts) are almost all due to baseline-docking failures on large, flexible molecules. All subsequent main results are computed over the 915 successfully completed compounds.

\begin{table}[t]
\centering
\caption{Main result over the full compound pool. Gap is defined in Eq.~\ref{eq:gap}; positive means target-preferring. The paired test compares each compound's own baseline against its own optimized result.}
\label{tab:main}
\begin{tabular}{lcc}
\toprule
& \textbf{Baseline (original drug)} & \textbf{After optimization} \\
\midrule
Mean specificity gap (kcal/mol)   & $-0.718$ & $\mathbf{+0.472}$ \\
Median specificity gap (kcal/mol) & $-0.600$ & $+0.300$ \\
Compounds improved                & --- & $776/915$ $(84.8\%)$ \\
Mean Tanimoto to original         & $1.000$ & $0.720$ \\
\bottomrule
\end{tabular}
\end{table}

\begin{figure}[t]
\centering
\includegraphics[width=\textwidth]{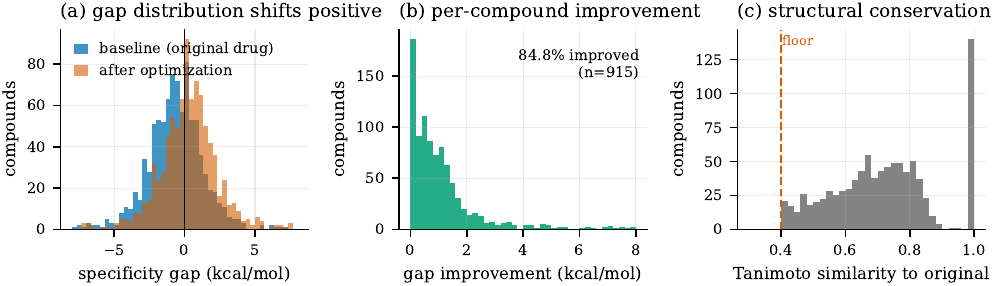}
\caption{\textbf{Main result across 915 compounds.} (a) The specificity-gap distribution shifts from predominantly negative (off-target-preferring) to predominantly positive. (b) Per-compound improvement. (c) Structural conservation; the spike at $1.0$ is compounds returned unchanged because no candidate passed the admission gates.}
\label{fig:main}
\end{figure}

Table~\ref{tab:main} and Figure~\ref{fig:main} summarize the overall results. On average, compounds shift from preferential binding to off-targets at baseline to preferential binding to their intended targets after optimization. Among the 575 compounds with a negative baseline specificity gap, 190 (33\%) achieve a positive gap after optimization. The similarity distribution (Figure~\ref{fig:main}c) shows that these improvements are achieved while largely preserving the original molecular structures. 

\subsection{Optimization examples and case studies}
\label{sec:cases}

Because SpecOpt records the contact-difference signal, LLM proposals, and accepted edits at each round, individual optimization trajectories are fully traceable. Figure~\ref{fig:worked} illustrates this process for edoxaban, a factor~Xa inhibitor with prothrombin as an off-target. The contact-difference analysis identifies the cyclohexane linker as a \textsc{grow} region, with three atoms having greater available pocket volume in factor~Xa, while two other fragments are marked for \textsc{redesign}. Consistent with this signal, the LLM proposes extending the cyclohexane ring. A methyl and fluorine modification are accepted in the first round, followed by addition of a trifluoromethyl group at the same position in the second round. Across these edits, the specificity gap improves from $-5.6$ to $+1.5$~kcal/mol while retaining a Tanimoto similarity of $0.76$ to the original compound.

\begin{figure}[t]
\small
\fbox{\parbox{0.98\textwidth}{
\textbf{Contact-difference summary supplied to the LLM (edoxaban, factor~Xa vs.\ prothrombin):}
\begin{itemize}[leftmargin=1.2em,itemsep=0pt,topsep=2pt]
\item \texttt{[1*][C@@H]1C[C@@H](C(=O)N(C)C)CC[C@@H]1[2*]}: contacts are shared with the off-target, but \textbf{3 atoms have substantially more available space in the target pocket}; candidate to \textsc{grow}.
\item \texttt{[2*]c1nc2c(s1)CN(C)CC2}: contacts occur in both poses at 7 atoms, but \textbf{against different residues}; candidate to \textsc{redesign}.
\item \texttt{[1*]C(=O)N[2*]}: contacts occur in both poses at 3 atoms, against different residues; candidate to \textsc{redesign}.
\item \texttt{[1*]Nc1ccc(Cl)cn1}, \texttt{[1*]NC(=O)C([2*])=O}: no docking contact data available.
\end{itemize}
\textbf{Accepted edits:}
\begin{itemize}[leftmargin=1.2em,itemsep=0pt,topsep=2pt]
\item Round 1: extend cyclohexane with methyl and fluorine modifications (gap $-5.6 \rightarrow -1.0$).
\item Round 2: add CF$_3$ at the same position (gap $-1.0 \rightarrow +1.5$).
\end{itemize}
\textbf{Final:} gap $-5.6 \rightarrow +1.5$~kcal/mol; Tanimoto similarity $0.76$.
}}
\caption{Example optimization trajectory for edoxaban. The contact-difference analysis identifies candidate regions for modification, and the LLM proposes corresponding structural edits that are evaluated and accumulated across optimization rounds.}
\label{fig:worked}
\end{figure}

Table~\ref{tab:cases} presents additional examples that begin with negative specificity gaps, achieve positive gaps after optimization, and retain Tanimoto similarity $\geq 0.6$ to the original compound. The examples span multiple target classes, including proteases, GPCRs, kinases, and metabolic enzymes. The resulting modifications are generally small, localized substitutions, including fluorine and CF$_3$ additions, nitrile addition, substituent repositioning, and methylation, consistent with the constrained lead-optimization setting of SpecOpt.

\begin{figure}[t]
\centering
\includegraphics[width=\textwidth]{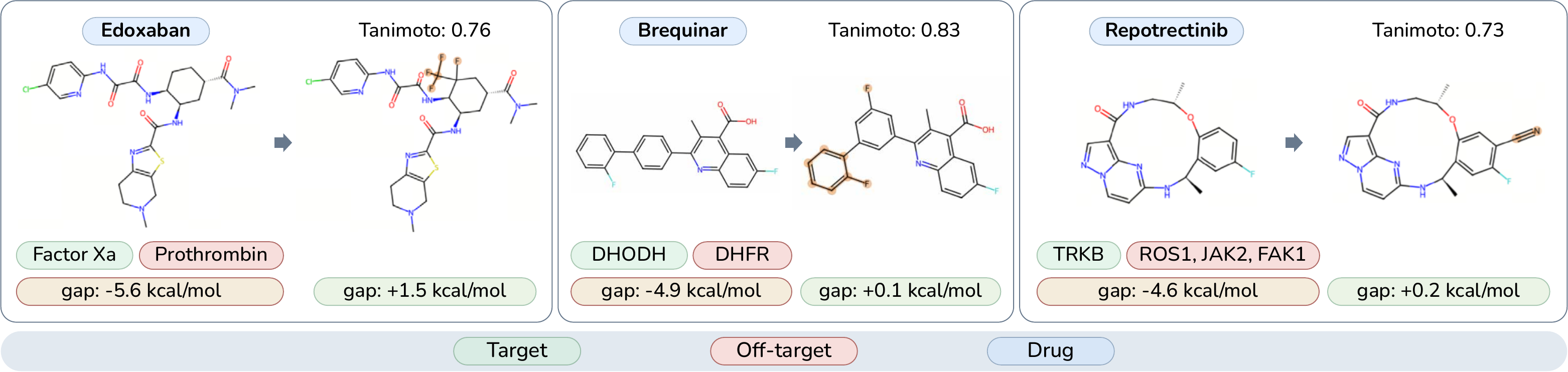}
\caption{Selected SpecOpt case studies showing the original (left) and optimized (right) structures. Highlighted atoms indicate structural changes relative to the maximum common substructure of the original compound. Specificity gaps are reported in kcal/mol.}
\label{fig:cases}
\end{figure}

\begin{table}[t]
\centering
\small
\caption{Selected case studies from the main evaluation. ``Edit'' summarizes the net structural change relative to the original compound.}
\label{tab:cases}
\begin{tabular}{llcccl}
\toprule
\textbf{Compound} & \textbf{Target / off-target} & \textbf{Base} & \textbf{Opt.} & \textbf{Sim.} & \textbf{Edit} \\
\midrule
Edoxaban & factor Xa / prothrombin & $-5.6$ & $+1.5$ & $0.76$ & F + CF$_3$ on cyclohexane \\
Ibipinabant & CB$_1$ / CB$_2$ & $-1.5$ & $+5.1$ & $0.73$ & F on arylsulfonyl ring \\
Brequinar & DHODH / DHFR & $-4.9$ & $+0.1$ & $0.83$ & F repositioned on biaryl \\
Repotrectinib & TRKB / ROS1, JAK2, FAK1 & $-4.6$ & $+0.2$ & $0.73$ & nitrile on fluoroaryl \\
Berotralstat & kallikrein / plasminogen & $-3.5$ & $+1.9$ & $0.78$ & NH $\rightarrow$ $N$-methyl \\
\bottomrule
\end{tabular}
\end{table}

\subsection{Ablation study: identifying the key signal}
\label{sec:ablation}

\begin{figure}[t]
\centering
\includegraphics[width=\textwidth]{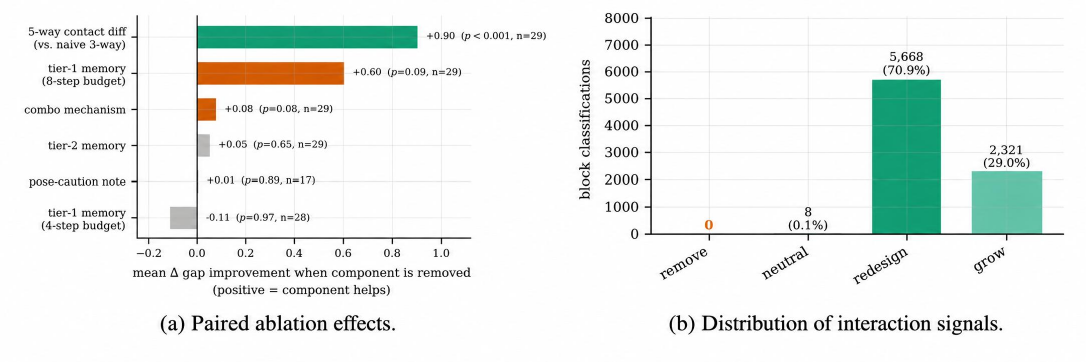}
\caption{Ablation and signal analysis. (a) Mean change in specificity-gap improvement when each component is removed; positive values indicate that the component improves performance. Green indicates $p<0.05$, orange $p<0.1$, and grey no statistical significance. (b) Fragment-level interaction signals aggregated across the 915-compound evaluation. Exclusive target- or off-target-contact signals do not occur, while most informative signals arise from different residue identities across shared contacts and target-specific growth opportunities.}
\label{fig:ablation}
\end{figure}

To isolate the contribution of each component, we fix a 30-compound subset stratified by baseline specificity gap and evaluate the full system against five ablated variants. Each ablation is compared with the full-system result on the same compound (Figure~\ref{fig:ablation}a).

\paragraph{Residue identity provides the critical optimization signal.} Replacing the residue-aware classification with a naive three-way classification (target-only, off-target-only, or shared contact) reduces the mean specificity-gap improvement from $+0.903$ to \textbf{$0.000$ across all 29/29 compounds}: without residue identity, the system detects no differentiating interaction signal and terminates without modification. Figure~\ref{fig:ablation}b explains this behavior. Across the full 915-compound evaluation, the dominant signals are \textsc{shared-different-residues} (5{,}668 fragment classifications) and \textsc{growth-opportunity} (2{,}321). Thus, simply identifying whether a ligand region contacts the target or off-target is insufficient; the useful signal comes primarily from differences in \emph{which residues} are contacted.

This observation also explains why the deterministic \textsc{remove} operation contributes no edits. The operation is designed to delete fragments that interact exclusively with an off-target, but its trigger condition does not occur in our evaluation. We therefore retain this operation in the framework but report its lack of activation rather than relaxing its definition post hoc.

\begin{table}[t]
\centering
\caption{Performance grouped by the relationship between the intended target and the off-target defining the specificity gap. Performance is consistent across relationship types.}
\label{tab:relation}
\begin{tabular}{lcccccc}
\toprule
\textbf{Relationship} & $n$ & \textbf{Base} & \textbf{Opt.} & $\Delta$ & \textbf{\% improved} & \textbf{Wilcoxon $p$} \\
\midrule
Isoform / subtype & 149 & $-0.72$ & $+0.48$ & $+1.20$ & $85.9\%$ & $9{\times}10^{-23}$ \\
Homologous (same class) & 335 & $-0.43$ & $+0.94$ & $+1.38$ & $83.3\%$ & $2{\times}10^{-47}$ \\
Different (no shared class) & 431 & $-0.94$ & $+0.10$ & $+1.04$ & $85.6\%$ & $3{\times}10^{-62}$ \\
\bottomrule
\end{tabular}
\end{table}

\paragraph{Mechanisms with no measurable performance benefit.} We evaluate three additional mechanisms that operate as intended but do not significantly improve optimization performance. (i)~\emph{Cross-compound memory} retrieves an average of 2.5 relevant records from previous compounds, but removing it has no significant effect on performance ($+0.052$, $p=0.65$). To test whether this is caused by limited relevance of the retrieved information, we repeat the experiment on 24 drugs with identical target and off-target sets, where cross-compound information should be maximally transferable, but again observe no benefit ($-0.067$, $p=0.33$). (ii)~The \emph{pose-consistency} mechanism identifies cases where a ligand adopts substantially different conformations in the target and off-target pockets and warns the LLM accordingly. We evaluate it on flexible molecules, where 10/17 compounds receive medium- or high-risk warnings, but observe no improvement ($-0.07$). (iii)~The \emph{combo} mechanism attempts to combine individually beneficial edits and activates for approximately 60\% of eligible compounds. Most combined edits (6/8) fail the ADMET acceptance criteria, and the overall improvement is small and not statistically significant ($+0.079$, $p=0.08$).

\paragraph{A budget-dependent effect.} Within-compound memory, which prevents the agent from repeating previously attempted edits, shows no measurable benefit under a 4-round optimization budget ($-0.111$, $p=0.97$). However, when the budget is increased to 8 rounds, memory produces a larger improvement ($+0.603$, $p=0.091$). The memory-enabled condition also evaluates more candidates on average ($27.3$ vs.\ $23.6$), suggesting that avoiding repeated proposals allows the optimization to continue exploring rather than triggering the ``no improvement'' early-stopping criterion. Because the main evaluation uses the 4-round budget, it may underestimate the benefit of within-compound memory for longer optimization runs.

\subsection{Multiple off-targets}
\label{sec:multioff}

For compounds with multiple off-targets, we retain up to three off-targets based first on structural availability and then on binding affinity. The contact-difference signals from these off-targets are pooled to construct a single fragment-level summary for the LLM, allowing SpecOpt to optimize against multiple competing proteins simultaneously. To examine the effect on a broader set of off-targets, we additionally re-dock the FLT3 benchmark compounds against five common off-targets (Appendix~\ref{app:multioff}, Figure~\ref{fig:multioff}). On average, optimization strengthens binding to the intended target while weakening binding to most off-targets. Only $5/20$ compounds weaken all five off-targets simultaneously, while others show mixed effects. This behavior is expected because the current implementation optimizes against at most three off-targets; proteins outside this selected set are not explicitly considered during optimization. Extending SpecOpt to jointly optimize against a larger off-target panel is a natural direction for future work.

\vspace{-0.4em}
\subsection{External comparison: FLT3}
\vspace{-0.4em}
\label{sec:flt3}

To compare SpecOpt with prior selectivity-aware molecular design, we adopt the setting of \citet{chandra2024bayesian}, using FLT3 as the intended target and PDGFRA, KIT, VEGFR2, MK2, and JAK2 as off-targets. Rather than optimizing newly generated molecules, we apply SpecOpt to 24 real FLT3-targeting drugs in our benchmark, including sunitinib, midostaurin, gilteritinib, and crenolanib. Among 23 successful runs, SpecOpt improves the specificity gap for 20 compounds, shifting the mean gap from $-2.03$ to $-0.91$~kcal/mol ($p=8.8\times10^{-5}$).

Absolute selectivity values are not directly comparable between the two methods because they address different tasks and use different docking pipelines. \citet{chandra2024bayesian} generate molecules \emph{de novo}, whereas SpecOpt modifies existing drugs under structural constraints. 

\vspace{-1em}
\section{Discussion and Limitations}
\vspace{-0.8em}

Our results highlight the importance of residue-aware differential contacts rather than contact presence alone for constrained molecular optimization. The LLM translates these signals into specific chemical edits, and richer descriptions of the local residue environment may further improve this process. Additional agentic mechanisms do not consistently improve performance, emphasizing the need to validate their contributions rather than assume that greater complexity is beneficial. Several limitations remain. First, SpecOpt optimizes against at most three selected off-targets. Evaluation against five off-targets shows that improvements can extend beyond this set, but some compounds also strengthen binding to individual off-targets; improved specificity against the selected proteins therefore does not guarantee broad selectivity. Second, all reported binding improvements are based on Vina docking scores. Fixed docking seeds support reproducible evaluation but do not establish experimental affinity or selectivity, which require experimental validation or assessment with higher-accuracy free-energy calculations.

\vspace{-1em}
\section{Conclusion}
\vspace{-0.8em}

We introduced \emph{specificity optimization} (SpecOpt), the task of modifying an existing drug to improve its binding preference for an intended target over known off-targets while preserving the original molecule. We constructed a benchmark for this task and developed an agentic framework combining differential structural analysis with LLM-guided molecular editing. Our results identify residue-aware differential contacts as the key optimization signal and demonstrate that existing drugs can be systematically modified toward improved predicted specificity while retaining substantial structural similarity. SpecOpt therefore provides a complementary direction to selective \emph{de novo} generation and a foundation for further development of specificity-aware lead optimization.

\subsubsection*{Reproducibility Statement}

The benchmark construction, filtering criteria, agent implementation, ablation toggles, and the exact per-compound outputs behind every number in this paper are contained in the accompanying code repository. All docking uses fixed seeds; the precomputed docking cache is released to make the main run reproducible without repeating $\sim$15k docking calls.
Code is available at 
\url{https://anonymous.4open.science/r/specopt-611E}.

\subsubsection*{Acknowledgments}
This work was supported by the NSF Molecule Maker Lab Institute (MMLI), an AI Institute for Molecular Discovery, Synthesis Strategy, and Manufacturing, funded by the U.S. National Science Foundation under Awards No. 2019897 and 2505932.


\bibliography{references}
\bibliographystyle{iclr2026_conference}

\clearpage
\appendix
\section{Appendix}

\subsection{Generalization across a broader off-target panel}
\label{app:multioff}

\begin{figure}[!hbt]
\centering
\includegraphics[width=\textwidth]{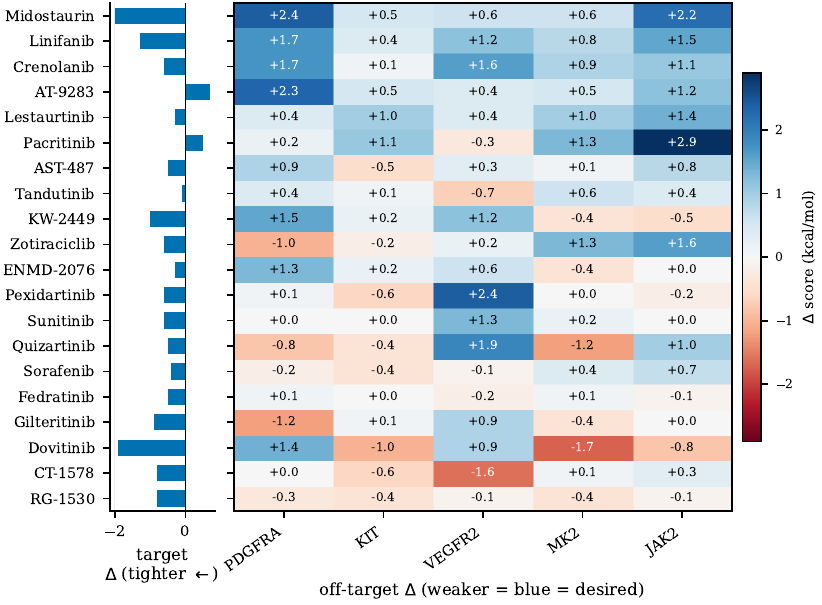}
\caption{Binding-score changes after optimization for FLT3-targeting compounds. Left: change in target binding, where negative values indicate stronger binding. Right: changes across five off-targets, where positive values (blue) indicate weaker binding. Five compounds weaken all five off-targets simultaneously, while the remaining compounds show mixed effects across off-targets.}
\label{fig:multioff}
\end{figure}

Figure~\ref{fig:multioff} examines whether optimization generalizes across a broader off-target panel. Most compounds show weaker predicted binding to multiple off-targets, and five compounds weaken all five simultaneously, indicating that improvements can generalize beyond individual proteins. However, several compounds strengthen binding to at least one off-target. This mixed behavior is expected because SpecOpt currently considers at most three off-targets during optimization, whereas the figure evaluates five; off-targets outside the selected optimization set are not explicitly constrained.

\subsection{Benchmark characterization and docking validation}
\label{app:eda}

We characterize the benchmark composition and assess the reliability and limitations of its docking scores.

\begin{figure}[t]
\centering
\includegraphics[width=\textwidth]{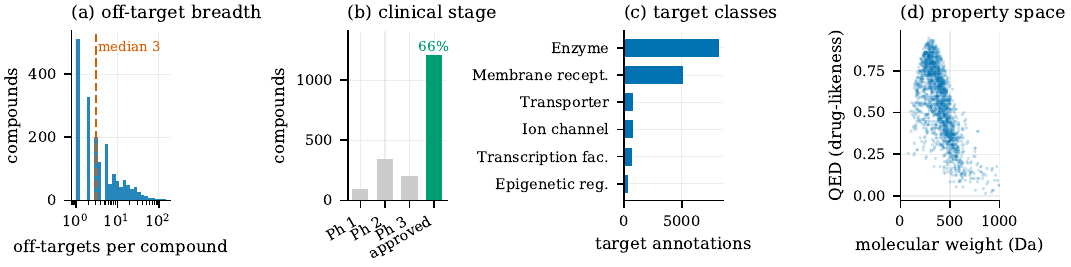}
\caption{Benchmark composition. (a)~Number of off-targets per compound on a logarithmic axis, with a median of 3. (b)~Clinical stage; approximately two thirds of compounds are approved drugs. (c)~Target-class annotations. (d)~Molecular weight versus QED for the starting compounds.}
\label{fig:eda_composition}
\end{figure}

\paragraph{Composition.} The benchmark contains 1{,}848 compounds spanning 994 distinct protein targets (Figure~\ref{fig:eda_composition}). The number of annotated off-targets per compound follows a heavy-tailed distribution, with a median of 3, a mean of 7.5, a 90th percentile of 18, and a maximum of 148. This variation is relevant to optimization because the specificity gap is defined by the tightest-binding off-target. Compounds with larger off-target sets may present more competing interactions, increasing the possibility that reducing binding to one off-target leaves another as the principal determinant of the specificity gap (Section~\ref{sec:multioff}).

The benchmark consists of drugs and drug candidates rather than generated molecules: 1{,}214 of 1{,}848 compounds (66\%) are approved (ChEMBL \texttt{max\_phase}~4), with the remainder in Phase~1--3. The median molecular weight is 383\,Da, and the median quantitative estimate of drug-likeness (QED)~\citep{bickerton2012qed} is 0.57. These baseline properties motivate the use of similarity and ADMET constraints to preserve characteristics of the starting compounds. Target annotations are dominated by enzymes (8{,}163) and membrane receptors (5{,}110), followed by transporters (785), ion channels (744), transcription factors, and epigenetic regulators. The benchmark therefore emphasizes kinases and GPCRs, and generalization to other target classes, such as protein--protein interaction targets, requires further evaluation.

\paragraph{Docking data.} We attempt to dock each compound against every protein annotated as an intended target or off-target. Receptors are resolved per UniProt~\citep{uniprot2025} accession, preferring an experimental PDB structure~\citep{berman2000pdb} over an AlphaFold model~\citep{jumper2021alphafold} and, among PDB candidates, the highest-resolution holo structure (one with a bound non-polymer ligand) at $\le 3.5$\,\AA; ligands are embedded with RDKit ETKDGv3~\citep{wang2020etkdg} and docked with AutoDock Vina. Both the conformer seed and the Vina seed are fixed to support reproducibility. We attempted 15{,}195 compound--protein pairs. 13{,}636 returned a score; of those, 143 (1.05\%) were physically implausible ($> -1$\,kcal/mol, 117 of them positive, maximum $+43.9$) and were discarded as failed poses, leaving \textbf{13{,}493 usable docking scores}: 1{,}670 against intended targets and 11{,}823 against off-targets. The remaining 1{,}559 pairs did not return a score, primarily because no protein structure could be resolved. The full cache is released with the benchmark.

\paragraph{Agreement with measured activity.} pChEMBL values can represent several endpoints ($K_i$, $K_d$, IC50, AC50, EC50, and Potency), so direct correlation with docking scores would combine binding constants with assay-dependent measures of activity. To reduce this heterogeneity, we queried ChEMBL at the \emph{activity} level for the 20 most frequent proteins in the benchmark and retained only binding assays (\texttt{assay\_type=B}), values reported in nM, and uncensored measurements (\texttt{standard\_relation} `='). For each protein, we selected the endpoint type with the greatest compound coverage, recomputed $p$-activity from the raw values, and summarized replicate measurements by their median. Coverage ranged from 89--99\% of compounds associated with each protein.

Table~\ref{tab:top5} reports the five proteins with the strongest correlations. Muscarinic~M2 achieves $\rho = -0.574$ across 132 compounds, and all five correlations are statistically significant and have the expected negative sign. These results indicate that docking scores provide information about relative activity for some well-characterized targets. However, these five proteins represent the strongest results among 19 evaluated proteins: the overall median is $\rho = -0.075$, and only 6/19 correlations reach $p<0.05$. Agreement is therefore target-dependent. Restricting the analysis to $K_i$ measurements yields negative correlations for 15/18 proteins, of which 7 are statistically significant, with a median correlation of $-0.092$.

\begin{table}[htbp]
\centering
\caption{The five proteins with the strongest docking--affinity agreement, after filtering to a single assay type and unit. Negative $\rho$ is the expected direction (a more negative Vina score should accompany a higher $p$-activity). These are the strongest correlations among the 19 proteins tested; the median over all 19 is $-0.075$.}
\label{tab:top5}
\begin{tabular}{llrrr}
\toprule
Protein & Endpoint & $n$ & Spearman $\rho$ & $p$ \\
\midrule
Muscarinic acetylcholine receptor M2 & AC50 & 132 & $-0.574$ & $6.3\times10^{-13}$ \\
Mu-type opioid receptor              & AC50 & 119 & $-0.316$ & $4.7\times10^{-4}$ \\
5-HT receptor 2B                     & Ki   &  99 & $-0.258$ & $0.010$ \\
KCNH2 (hERG)                         & IC50 & 124 & $-0.237$ & $0.008$ \\
5-HT receptor 2A                     & AC50 & 117 & $-0.229$ & $0.013$ \\
\bottomrule
\end{tabular}
\end{table}

\begin{figure}[h]
\centering
\includegraphics[width=0.72\textwidth]{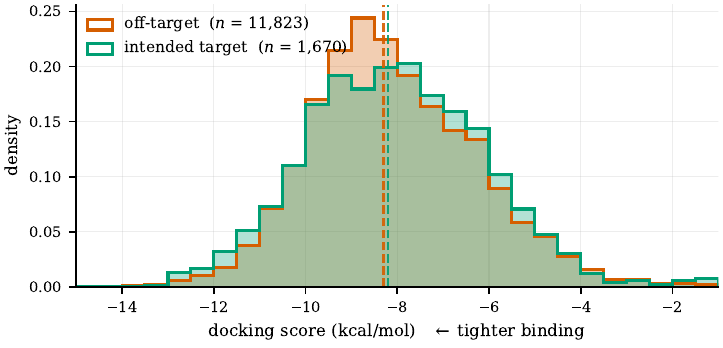}
\caption{Docking scores pooled into two sets over the whole benchmark: compound--protein pairs where the protein is the compound's intended target ($n = 1{,}670$) and bindings where it is an off-target ($n = 11{,}823$). Density-normalized; dashed lines mark medians. The comparison shows no statistically significant difference and a small effect size: medians $-8.20$ vs.\ $-8.30$\,kcal/mol, Mann--Whitney $p = 0.21$, Cohen's $d = 0.02$, with quantiles agreeing to within $0.1$--$0.4$\,kcal/mol at the 5th, 25th, 75th and 95th percentiles.}
\label{fig:eda_dist}
\end{figure}

\paragraph{Intended-target and off-target score distributions.} Figure~\ref{fig:eda_dist} groups all usable docking scores according to whether the protein is an intended target or an off-target of the compound. Both distributions are unimodal, centered near $-8.2$\,kcal/mol, and span approximately $-14$ to $-2$\,kcal/mol. The two distributions are closely aligned.

\paragraph{Binding-strength categories.} We interpret Vina scores as approximate binding free energies to define nominal dissociation-constant thresholds using $\Delta G = RT \ln K_d$ ($RT = 0.593$\,kcal/mol at 298\,K). Table~\ref{tab:bindcat} groups all usable docking scores by these thresholds and categorizes experimental measurements using the corresponding pChEMBL cutoffs.

\begin{table}[htbp]
\centering
\caption{Binding-strength categories. Left: our docking scores. Right: the experimental affinities binned at the pChEMBL equivalents of the same cut points ($-9.0 \to 6.59$, $-7.0 \to 5.13$, $-5.0 \to 3.66$). Docking places intended targets and off-targets in nearly identical proportions; the experimental labels differ by approximately 54 percentage points in the strong category. The experimental ``negligible'' row is empty because off-targets enter the benchmark only at pChEMBL $\ge 5.0$.}
\label{tab:bindcat}
\small
\begin{tabular}{llrrcrr}
\toprule
& & \multicolumn{2}{c}{Docking score} & & \multicolumn{2}{c}{Experimental (pChEMBL)} \\
\cmidrule(lr){3-4}\cmidrule(lr){6-7}
Category & Vina / $K_d$ & target & off-target & & target & off-target \\
\midrule
Strong     & $\le -9.0$ \ ($\lesssim$ 260\,nM)       & 34.7\% & 34.5\% & & 81.5\% & 27.7\% \\
Moderate   & $-9.0$ to $-7.0$ \ (260\,nM--7.5\,\textmu M) & 37.6\% & 40.2\% & & 15.4\% & 64.1\% \\
Weak       & $-7.0$ to $-5.0$ \ (7.5--220\,\textmu M)     & 22.6\% & 20.4\% & & \phantom{0}3.1\% & \phantom{0}8.2\% \\
Negligible & $> -5.0$ \ ($\gtrsim$ 220\,\textmu M)        & \phantom{0}5.1\% & \phantom{0}4.9\% & & \phantom{0}0.0\% & \phantom{0}0.0\% \\
\midrule
\multicolumn{2}{l}{Strong or moderate} & 72.3\% & 74.7\% & & 96.9\% & 91.8\% \\
\multicolumn{2}{l}{$n$} & 1{,}670 & 11{,}823 & & 1{,}911 & 13{,}813 \\
\bottomrule
\end{tabular}
\end{table}

These results illustrate both the utility and the limitations of the docking scores. Under the thresholds in Table~\ref{tab:bindcat}, 74.6\% of compound--protein pairs fall within the strong or moderate categories, while 4.9\% fall within the negligible category. Although this distribution does not independently establish predictive accuracy, the target-specific correlations in Table~\ref{tab:top5} support the use of docking scores for relative comparisons in some settings.

However, the scores provide limited separation between intended targets and off-targets. Docking assigns similar proportions to the strong or moderate categories (72.3\% and 74.7\%, respectively), with differences of at most 2.6 percentage points in any category. In contrast, experimental measurements place 81.5\% of intended targets and 27.7\% of off-targets in the strong category, a difference of approximately 54 percentage points; the corresponding docking difference is only 0.2 percentage points. Absolute Vina scores therefore do not reliably distinguish intended targets from off-targets in this benchmark. Our optimization evaluation instead uses pre- and post-modification \emph{differences} computed with fixed receptors and seeds. This comparison cancels systematic offsets to the extent that they are shared between the two evaluations, but does not eliminate all scoring errors.

\subsection{Docking setup}
\label{app:vina}

All docking uses AutoDock Vina with receptors and ligands prepared using MGLTools/AutoDockTools~\citep{morris2009autodocktools}. We report the configuration and implementation safeguards used to address failures observed during large-scale docking.

\paragraph{Receptor selection.} ChEMBL target identifiers are mapped to UniProt accessions and then to structures through RCSB PDB~\citep{berman2000pdb}. We query X-ray or cryo-EM entries with resolution $\le 3.5$\,\AA\ and prefer a holo structure (one containing at least one bound non-polymer entity) over an apo structure with marginally higher resolution. If no holo candidate is available, we select the highest-resolution structure. For example, ranking FLT3 structures solely by resolution selects the apo entry 1RJB (2.10\,\AA) over the holo entry 6JQR, despite a resolution difference of only $0.1$\,\AA. Prioritizing holo structures accounts for differences in binding-pocket geometry between ligand-bound and unbound conformations.

\paragraph{Search box.} The search box is defined using the following procedures, in order of priority.

\begin{enumerate}[leftmargin=1.6em,itemsep=1pt,topsep=2pt]
\item \textbf{Co-crystal ligand present.} The box is centered on the bound ligand and sized to its coordinate span plus 12\,\AA\ of padding. Coordinates are taken from the \emph{first occurrence} of the ligand rather than pooled across the structure. In homo-oligomeric structures, pooling ligand copies from multiple subunits can expand a pocket-centered box to encompass the full assembly. For ALDH2/1O04, this produced a $53\times161\times131$\,\AA\ box, exceeding the volume of a typical $\sim$20\,\AA\ pocket box by more than 100-fold, and increased Vina runtime to 475\,s compared with a typical 10--20\,s.
\item \textbf{No co-crystal ligand.} We apply \texttt{fpocket}~\citep{leguilloux2009fpocket} to the protein-only receptor and define the box around the highest-ranked pocket with 12\,\AA\ of padding. Pockets are ranked by fpocket's \emph{druggability} score~\citep{schmidtke2010druggability} rather than its default geometric score. These rankings can differ: for 2ATX, the default first-ranked pocket had a druggability score of 0.379, whereas the third-ranked pocket scored 0.853. If no suitable pocket is identified, the fallback is a whole-protein search box with 10\,\AA\ of padding.
\end{enumerate}

\paragraph{Vina parameters.} \texttt{--exhaustiveness 8}, \texttt{--cpu 1}, \texttt{--seed 42}; the reported score is the top-ranked mode from the Vina log. Ligands are embedded with RDKit ETKDGv3~\citep{wang2020etkdg} and optimized with MMFF~\citep{halgren1996mmff} before conversion to PDBQT.

\paragraph{Determinism.} Our reproducibility configuration fixes three parameters: the Vina seed, the Vina thread count (\texttt{--cpu 1}), and the RDKit conformer seed. The single-thread setting avoids variation associated with thread scheduling. RDKit's \texttt{EmbedMolecule} uses a default \texttt{randomSeed} of $-1$, so failing to specify this seed can produce different starting conformers even when the Vina seed is fixed. Before fixing these parameters, repeated docking of granisetron against the same target produced different binding modes and scores. With all three parameters fixed, repeated runs produced byte-identical cache contents.

\paragraph{Concurrency.} By default, Vina detects all available CPUs (128 on our system). Concurrent docking calls can therefore oversubscribe computational resources. With six simultaneous processes, we observed SIGSEGV failures and nonzero exit codes that did not occur when the same commands were executed individually. We consequently restrict each Vina invocation to one thread and parallelize across independent docking processes. Each call uses a separate temporary directory to prevent concurrent processes from overwriting intermediate files and corrupting results.

\paragraph{Timeouts.} Each subprocess call has an explicit timeout: 300\,s for Vina and 120\,s for MGLTools. The Python~2 script \texttt{prepare\_ligand4.py} can fail to terminate for certain ligand topologies; in two observed cases, calls continued at 99.9\% CPU utilization for 70 minutes without producing output. Because \texttt{subprocess.run} has no default timeout, explicit limits are necessary to prevent individual ligand-processing failures from indefinitely occupying workers and progressively reducing available processing capacity.

\paragraph{Contacts.} For contact-difference analysis, we extract per-ligand-atom contacts using a 7.0\,\AA\ threshold and set \texttt{max\_hotspots} to 150. The underlying routine was designed for fragment growth and prioritizes a small number of high-volume sites. We increase the default limit of 5 to obtain broader coverage of the ligand--protein interface.

\paragraph{Caching.} Scores are cached for each (compound, protein) pair in a JSON file shared across the docking campaign. Writes use a temporary file followed by atomic replacement with \texttt{os.replace}. Readers retry after incomplete reads and retain the most recent valid copy. These safeguards address \texttt{JSONDecodeError} failures observed when concurrent processes accessed partially written cache files. The cache is released with the benchmark to enable reproduction of the main analysis without repeating approximately 15{,}000 docking calls.

\subsection{Ablation detail}

Table~\ref{tab:ablation} reports the numerical results corresponding to Figure~\ref{fig:ablation}a. Each row compares the full system with one ablated variant using paired observations from the same compounds.

\begin{table}[htbp]
\centering
\caption{Paired ablation results. $\Delta$ is mean(full $-$ ablated); positive means the component contributes.}
\label{tab:ablation}
\begin{tabular}{@{}lccccl@{}}
\toprule
\textbf{Component removed} & $n$ & \textbf{full} & \textbf{ablated} & $\Delta$ & \textbf{Wilcoxon $p$} \\
\midrule
Five-way contact diff (naive 3-way instead) & 29 & $+0.903$ & $+0.000$ & $\mathbf{+0.903}$ & $<0.001$ \\
Within-compound memory (8-round budget)     & 29 & $+1.390$ & $+0.786$ & $+0.603$ & $0.091$ \\
Combo mechanism                             & 29 & $+0.903$ & $+0.824$ & $+0.079$ & $0.082$ \\
Cross-compound memory                       & 29 & $+0.903$ & $+0.852$ & $+0.052$ & $0.648$ \\
Pose-consistency caution                    & 17 & $+1.147$ & $+1.141$ & $+0.006$ & $0.893$ \\
Within-compound memory (4-round budget)     & 28 & $+0.936$ & $+1.046$ & $-0.111$ & $0.972$ \\
\bottomrule
\end{tabular}
\end{table}

\subsection{ADMET comparison on the FLT3 setting}

Using the same six-endpoint ADMET oracle, we compare the optimized molecules with both their starting compounds and the reference method's generated candidates (Table~\ref{tab:admet}). The optimized molecules have lower predicted AMES, P-gp, and DILI probabilities and higher predicted HIA than both comparison groups. In particular, the mean DILI probability is $0.745$, compared with $0.753$ for the starting compounds and $0.787$ for the generated candidates. The improvements are not uniform across endpoints: predicted hERG probability increases slightly from $0.696$ to $0.702$, while BBBP decreases from $0.680$ to $0.619$ but remains above the generated candidates' $0.387$. These results indicate favorable changes in several predicted properties alongside endpoint-specific trade-offs. Because the methods start from different molecular populations and these scores are model predictions rather than experimental measurements, the comparison does not establish overall pharmacological superiority.

\begin{table}[htbp]
\centering
\caption{ADMET comparison (mean predicted probability). Arrows indicate the preferred direction: $\downarrow$ lower is better; $\uparrow$ higher is better.}
\label{tab:admet}
\begin{tabular}{@{}lccc@{}}
\toprule
\textbf{Endpoint} & \textbf{Generated (top-20)} & \textbf{Real drugs (baseline)} & \textbf{Ours (optimized)} \\
\midrule
AMES (mutagenicity) $\downarrow$   & $0.539$ & $0.288$ & $\mathbf{0.178}$ \\
hERG (cardiotoxicity) $\downarrow$ & $\mathbf{0.691}$ & $0.696$ & $0.702$ \\
P-gp (efflux) $\downarrow$         & $0.349$ & $0.315$ & $\mathbf{0.290}$ \\
DILI (liver injury) $\downarrow$   & $0.787$ & $0.753$ & $\mathbf{0.745}$ \\
HIA (absorption) $\uparrow$ & $0.442$ & $0.857$ & $\mathbf{0.876}$ \\
BBBP (BBB penetration) & $0.387$ & $0.680$ & $0.619$ \\
\bottomrule
\end{tabular}
\end{table}

\subsection{Implementation details}

The complete docking configuration is provided in Section~\ref{app:vina}. Molecule handling uses RDKit~\citep{landrum_rdkit}, and fragmentation follows BRICS~\citep{degen2008brics}. The minimum Tanimoto similarity is 0.4 relative to the original molecule, rather than the molecule from the preceding round, to constrain cumulative structural deviation. The ADMET tolerance is 0.15 for AMES, DILI, hERG, and P-gp. Unless otherwise specified, runs use 8 proposals per round and 4 rounds.

\end{document}